\documentclass[runningheads]{llncs}
\usepackage{graphicx}
\usepackage{comment}
\usepackage{amsmath,amssymb,amsfonts} 
\usepackage{color}
\usepackage{times}
\usepackage{epsfig}
\usepackage{enumitem}
\usepackage{booktabs}
\usepackage{multirow}
\usepackage{graphicx}
\usepackage{caption}
\usepackage{mathtools}
\usepackage{url}
\usepackage{subfiles}
\usepackage{bbold}
\usepackage[dvipsnames]{xcolor}
\usepackage{floatrow}
\usepackage{microtype}
\usepackage{booktabs}
\usepackage{colortbl}
\usepackage{diagbox}
\usepackage{caption}
\usepackage{subcaption}
\usepackage{floatrow}
\newfloatcommand{capbtabbox}{table}[][\FBwidth]

\usepackage{blindtext}

\def\eqnvspace{{\vspace{-2mm}}}
\def\figvspace{{\vspace{-3.1mm}}}

\definecolor{grey}{rgb}{0.1,0.1,0.1}
\DeclareMathOperator*{\argmin}{arg\,min} 
\newcommand{\Paragraph}[1]{\vspace{-0.03mm}\noindent\textbf{#1}\hspace{0mm}}
\newcommand{\Section}[1]{\vspace{-2mm}\section{#1}\hspace{-1mm}}
\newcommand{\SubSection}[1]{\vspace{-2.2mm}\subsection{#1}\hspace{-1mm}}

\begin{document}
\pagestyle{headings}
\mainmatter
\def\ECCVSubNumber{3004} 

\title{On Disentangling Spoof Trace for Generic Face Anti-Spoofing}

\titlerunning{On Disentangling Spoof Trace for Generic Face Anti-Spoofing}
\author{Yaojie Liu \and
Joel Stehouwer \and
Xiaoming Liu}
\authorrunning{Y. Liu et al.}
\institute{Michigan State University, East Lansing MI 48823, USA\\
\email{\{liuyaoj1,stehouw7,liuxm\}@msu.edu}}
\maketitle
\begin{center}\centering
    \includegraphics[width=\textwidth]{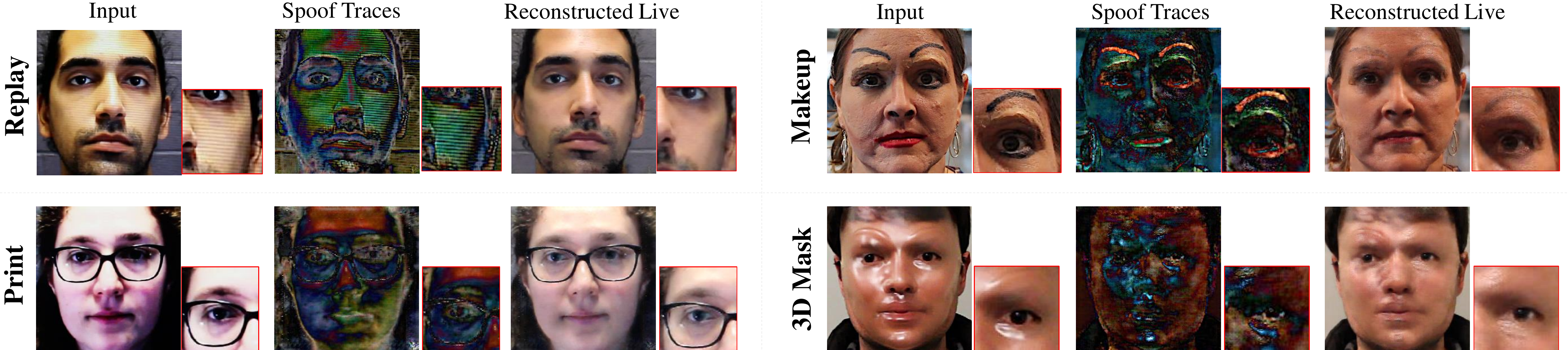}
    \captionof{figure}{\small The proposed approach can detect spoof faces, disentangle the spoof traces, and reconstruct the live counterparts. It can be applied to various spoof types and recognize diverse traces ({\it e.g.}, Moiré pattern in replay attack, artificial eyebrow and wax in makeup attack, color distortion in print attack, and specular highlights in $3$D mask attack). Zoom in for details.}
    \label{fig:1}
\figvspace
\end{center}%
\begin{abstract}

Prior studies show that the key to face anti-spoofing lies in the subtle image pattern, termed ``spoof trace", \textit{e.g.}, color distortion, $3$D mask edge, Moiré pattern, and many others. 
Designing a generic anti-spoofing model to estimate those spoof traces can improve not only the generalization of the spoof detection, but also the interpretability of the model's decision.
Yet, this is a challenging task due to the diversity of spoof types and the lack of ground truth in spoof traces. 
This work designs a novel adversarial learning framework to disentangle the spoof traces from input faces as a hierarchical combination of patterns at multiple scales.
With the disentangled spoof traces, we unveil the live counterpart of the original spoof face, and further synthesize realistic new spoof faces after a proper geometric correction.
Our method demonstrates superior spoof detection performance on both seen and unseen spoof scenarios while providing visually-convincing estimation of spoof traces.
Code is available at \url{https://github.com/yaojieliu/ECCV20-STDN}.

\end{abstract}
\Section{Introduction}
\label{sec:intro}
In recent years, the vulnerability of face biometric systems has been widely recognized and brought increasing attention to the vision community due to various physical and digital attacks. There are various physical and digital attacks, such as face morphing~\cite{dale2011video,zakharov2019few,zollhofer2018state}, face adversarial attacks~\cite{deb2019advfaces,goodfellow2014explaining,szegedy2013intriguing}, face manipulation attacks (\textit{e.g.}, deepfake, face swap)~\cite{deepfake,thies2016face2face}, and face spoofing (\textit{i.e.}, presentation attacks)~\cite{bigun2004assuring,frischholz2003avoiding,schuckers2002spoofing}, that can be used to attack the biometric systems.
Among all these attacks, face spoofing is the only physical attack to deceive the systems, where attackers present faces from spoof mediums, such as photograph, screen, mask and makeup, instead of a live human.
These spoof mediums can be easily manufactured by ordinary people, therefore posing a huge threat to applications such as mobile face unlock, building access control, and transportation security.
Therefore, face biometric systems need to be reinforced with face anti-spoofing techniques before performing face recognition tasks.

Face anti-spoofing\footnote{As most face recognition systems are based on a monocular camera, this work only concerns monocular face anti-spoofing methods, and terms as face anti-spoofing hereafter for simplicity.} has been studied for over a decade, and one of the most common approaches is based on texture analysis~\cite{boulkenafet2015face,boulkenafet2016face,patel2016secure}.
Researchers noticed that presenting faces from spoof mediums introduces special texture differences, such as color distortions, unnatural specular highlights, Moiré patterns and \textit{etc}.
Those texture differences are inherent within spoof mediums and thus hard to remove or camouflage.
Early works build a conventional feature extractor plus classifier pipeline, such as LBP+SVM and HOG+SVM~\cite{de2012lbp,komulainen2013context}.
Recent works leverage deep learning techniques and show great progress~\cite{atoum-depth-fas,liu-auxiliary-fas,deep-tree,shao2019multi,yang2019face}.

However, there are two limitations in the deep learning-based approaches. 
First, most prior works concern limited spoof types, either print/replay or $3$D mask alone, while a real-world anti-spoofing system may encounter a wide variety of spoof types including print, replay, $3$D masks, and facial makeup. 
Second, many approaches formulate face anti-spoofing as merely a classification/regression problem, with a single score as the output. 
Although a few methods~\cite{liu-auxiliary-fas,jourabloo-face-despoofing,yang2019face} attempt to offer insights via fixation, saliency, or noise analysis, there is little understanding on what the exact differences are between live and spoof, and what patterns the classifier's decision is based upon. 

We regard the face spoof detection for {\it all} existing spoof types as \textbf{generic face anti-spoofing}, and term the patterns differentiating a spoof face and its live counterpart as \textbf{spoof trace}. 
As shown in Fig.~\ref{fig:1}, this work aims to equip generic face anti-spoofing models with the ability to explicitly extract the spoof traces from the input faces. 
We term this process as \textbf{spoof trace disentanglement}.
This is a challenging objective due to the diversity of spoof traces and the lack of ground truth of traces. However, we believe that tackling this problem can bring several benefits:
\vspace{-1mm}
\begin{enumerate}
    \item Binary classification for face anti-spoofing would harvest any cue that helps classification, which might include spoof-irrelevant cues such as lighting, and thus hinder generalization. In contrast, spoof trace disentanglement explicitly tackles the most fundamental cue in spoofing, upon which the classification can be grounded and witnesses better generalization.
    \item With the trend of pursuing explainable AI~\cite{darpa-xai,arrieta2020explainable}, it is desirable for the face anti-spoofing model to generate the spoof patterns that support its binary decision,
    and spoof trace serves as a good visual explanation of the model's decision. Certain properties (\textit{e.g.}, severity, methodology) of spoof attacks could potentially be revealed based on the traces.
    \item Spoof traces are good sources for synthesizing realistic spoof samples. High-quality synthesis can address the issue of limited training data for the minority spoof types, such as special $3$D masks and makeup. 
\end{enumerate}

\begin{figure*}[t]
    \centering
    \resizebox{1\linewidth}{!}{\includegraphics{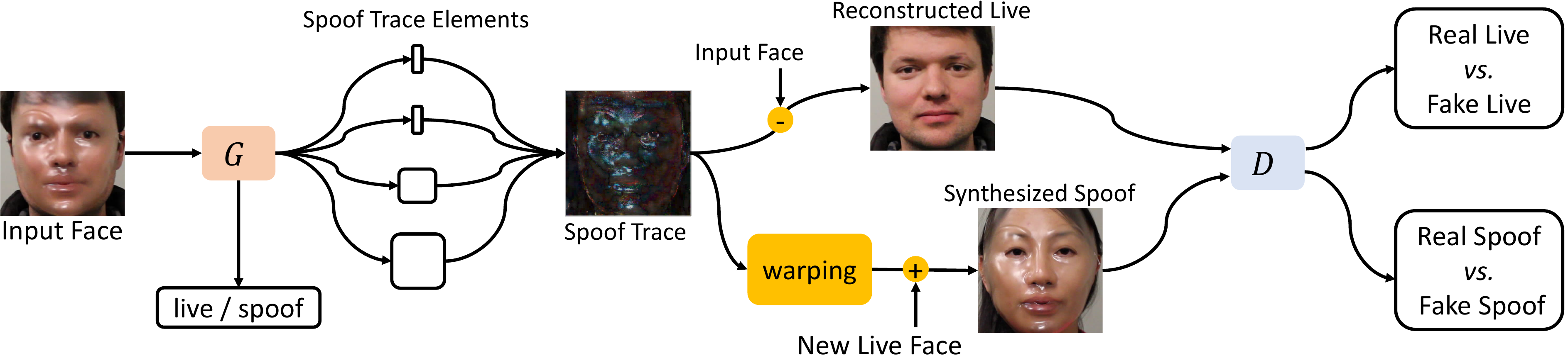}}
    \caption{\small Overview of the proposed Spoof Trace Disentanglement Network (STDN).}
    \label{fig:2}
\figvspace
\end{figure*}

As shown in Fig.~\ref{fig:2}, we propose a Spoof Trace Disentanglement Network (STDN) to tackle this problem.
Given only the binary labels of live \textit{vs.}~spoof, STDN adopts an overall GAN training strategy.
The generator takes input faces, detect the spoof faces, and disentangles the spoof traces as the combination of multiple elements.
With the spoof traces, we can reconstruct the live counterpart from the spoof and synthesize new spoof from the live.
To correct possible geometric discrepancy during spoof synthesis, we propose a novel $3$D warping layer to deform spoof traces toward the target face.
We deploy multiscale discriminators to improve the fidelity of both the reconstructed live and synthesized spoof. 
Moreover, the synthesized spoof samples are further utilized to train the generator in a supervised fashion, thanks to disentangled spoof traces as ground truth for the synthesized sample.

In summary, the main contributions of this work are as follows:
\vspace{-1mm}
\begin{itemize}
\item[$\bullet$] We for the first time study spoof trace for generic face anti-spoofing; 
\item[$\bullet$] We propose a novel model to disentangle spoof traces into a hierarchical representation;
\item[$\bullet$] We utilize the spoof traces to synthesize new data and enhance the training; 
\item[$\bullet$] We achieve SOTA anti-spoofing performance and provide convincing visualization. 
\end{itemize}

\Section{Related Work}
\label{sec:prior}
\Paragraph{Face Anti-Spoofing}:
Face anti-spoofing has been studied for more than a decade and its development can be roughly divided into three stages.
In early years, researchers leverage the spontaneous human movement, such as eye blinking and head motion, to detect simple print photograph or static replay attacks~\cite{kollreider2007real,pan2007eyeblink}. 
However, when facing counter attacks, such as print face with eye region cut, and replaying a face video, those methods would  fail.
Later, researchers pay more attention to texture differences between live and spoof, which are inherent with spoof mediums.
Researchers mainly extract handcrafted features from the faces, \textit{e.g.}, LBP~\cite{boulkenafet2015face,de2012lbp,de2013can,maatta2011face}, HoG ~\cite{komulainen2013context,yang2013face}, SIFT~\cite{patel2016secure} and SURF~\cite{boulkenafet2016face}, and train a classifier to discern the live \textit{vs.}~spoof, such as SVM and LDA.
Recently, face anti-spoofing solutions equip with deep learning techniques and demonstrate significant improvements over the conventional methods. 
Methods in~\cite{feng2016integration,li2016original,patel2016cross,yang2014learn} train a deep neural network to learn a binary classifier between live and spoof.
In~\cite{atoum-depth-fas,liu-auxiliary-fas,deep-tree,shao2019multi,yang2019face}, additional supervisions, such as face depth map and rPPG signal, are proposed to help the network to learn more generalizable features.
With the latest approaches achieving saturated performance on several benchmarks, researchers start to explore more challenging cases, such as few-shot/zero-shot face anti-spoofing~\cite{deep-tree,qin2019learning,zhao2019meta}, domain adaptation in face anti-spoofing~\cite{shao2019multi,shao2019regularized}, \textit{etc}.

In this work, we aim to solve an interesting but very challenging problem: disentangling and visualizing the spoof traces from the input faces. 
Related works~\cite{jourabloo-face-despoofing,stehouwer2020noise,camera-trace-erasing} also adopt GAN seeking to estimate the different traces. However, they formulate the traces as low-intensity noises, which is limited to print and replay attacks and cannot provide convincing visual results.
In contrast, we explore spoof traces from  a wide range of spoof attacks, visualize them with novel disentanglement, and also evaluate the proposed method on the challenging cases (\textit{e.g.}, zero-shot face anti-spoofing).

\Paragraph{Disentanglement Learning}: Disentanglement learning is often adopted to better represent complex data and features.
DR-GAN~\cite{disentangled-representation-learning-gan-for-pose-invariant-face-recognition,representation-learning-by-rotating-your-faces} disentangles face into identity and pose vectors for pose-invariant face recognition and view synthesis.
Similarly in gait recognition, ~\cite{gait-recognition-via-disentangled-representation-learning} disentangles the representations of appearance, canonical, and pose features from an input gait video.
$3$D reconstruction works~\cite{disentangling-features-in-3d-face-shapes-for-joint-face-reconstruction-and-recognition} also disentangle the representation of a $3$D face into identity, expressions, poses, albedo, and illuminations.
To solve the problem of image synthesis, \cite{esser2018variational} disentangles an image into appearance and shape with U-Net and Variational Auto Encoder (VAE).
Different from~\cite{disentangling-features-in-3d-face-shapes-for-joint-face-reconstruction-and-recognition,disentangled-representation-learning-gan-for-pose-invariant-face-recognition,gait-recognition-via-disentangled-representation-learning}, we intend to disentangle features that have different scales and contain geometric information. We leverage the multiple outputs from different layers to represent features at different scales, and adopt multiple-scale discriminators to properly learn them. Moreover, we propose a novel warping layer to handle the geometric information during the disentanglement and reconstruction.

\Section{Spoof Trace Disentanglement Network}\label{sec:method}
\vspace{-5mm}
\SubSection{Problem Formulation}
\label{sec:3-1}
Let the domain of live faces be denoted as $\mathcal{L}\! \subset \! \mathbb{R}^{N\!\times \!N \!\times \!3}$ and spoof faces as $\mathcal{S}\! \subset \!\mathbb{R}^{N\!\times \!N\! \times \!3}$, where $N$ is the image size.
We intend to obtain not only the correct prediction (live \textit{vs.} spoof) of the input face, but also a convincing estimation of the spoof traces.
Without the guidance of ground truth spoof traces, our key idea is to find a minimum change that transfers an input face to the live domain:
\eqnvspace
\begin{equation}
\label{eq:op}
\argmin_{\hat{\textbf{I}}} \| \textbf{I}-\hat{\textbf{I}}\|_F \; s.t. \; \textbf{I} \in (\mathcal{S}\cup\mathcal{L}) \; \text{and} \; \hat{\textbf{I}} \in \mathcal{L},
\end{equation}
where $\textbf{I}$ is the input face from either domain, $\hat{\textbf{I}}$ is the target face in the live domain, and $\textbf{I}-\hat{\textbf{I}}$ is defined as the spoof trace.
For an input live face $\textbf{I}_{\text{live}}$, the spoof traces should be $0$ as it's already in $\mathcal{L}$. For an input spoof face $\textbf{I}_{\text{spoof}}$, this $L$-$2$ regularization on spoof traces is also preferred, as there is no paired solution for the domain transfer and we hope the spoof traces to be bounded.
Based on~\cite{jourabloo-face-despoofing,patel2016secure}, spoof traces can be partitioned into multiple elements based on scales: global traces, low-level traces, and high-level traces.
Global traces, such as color balance bias and range bias, can be efficiently modeled by a single value.
The color biases here only refer to those created by the interaction between spoof mediums and the capturing camera, and the model is expected to ignore those spoof-irrelevant color variations.
Low-level traces consist of smooth content patterns, such as makeup strokes, and specular highlights.
High-level traces include sharp patterns and high-frequency texture, such as mask edges and Moiré pattern.
Denoted as $G(\cdot)$, the spoof trace disentanglement is formulated as a coarse-to-fine spoof effect build-up: 
\eqnvspace
\begin{equation}
\label{eq:spoof}
\begin{aligned}
G(\textbf{I}) &= \textbf{I} - \hat{\textbf{I}} \\
&= \textbf{I} - ((1-\textbf{s})\textbf{I} - \textbf{b} - \lfloor\textbf{C}\rfloor_N- \mathbf{T}) \\
&= \textbf{s}\textbf{I} + \textbf{b} + \lfloor\textbf{C}\rfloor_N + \mathbf{T},
\end{aligned}
\end{equation}
where $\textbf{s},\textbf{b}\in\mathbb{R}^{1\!\times \!1\!\times \!3} $ represent color range bias and balance bias, $\textbf{C} \in \mathbb{R}^{L\!\times \!L\!\times \!3}$ denotes the smooth content patterns ($L\!\!<\!\!N$ to enforce the smoothness), $\lfloor\cdot\rfloor$ is the resizing operation, and $ \textbf{T} \in \mathbb{R}^{N\!\times \!N\! \times \!3}$ is the high-level texture patterns.
Compared to the single layer representation~\cite{jourabloo-face-despoofing}, this disentangled representation $\{\textbf{s},\textbf{b},\textbf{C},\textbf{T}\}$ can largely improve disentanglement quality and suppress unwanted artifacts due to its coarse-to-fine process.

\begin{figure*}[t!]
    \centering
    \resizebox{\linewidth}{!}{\includegraphics{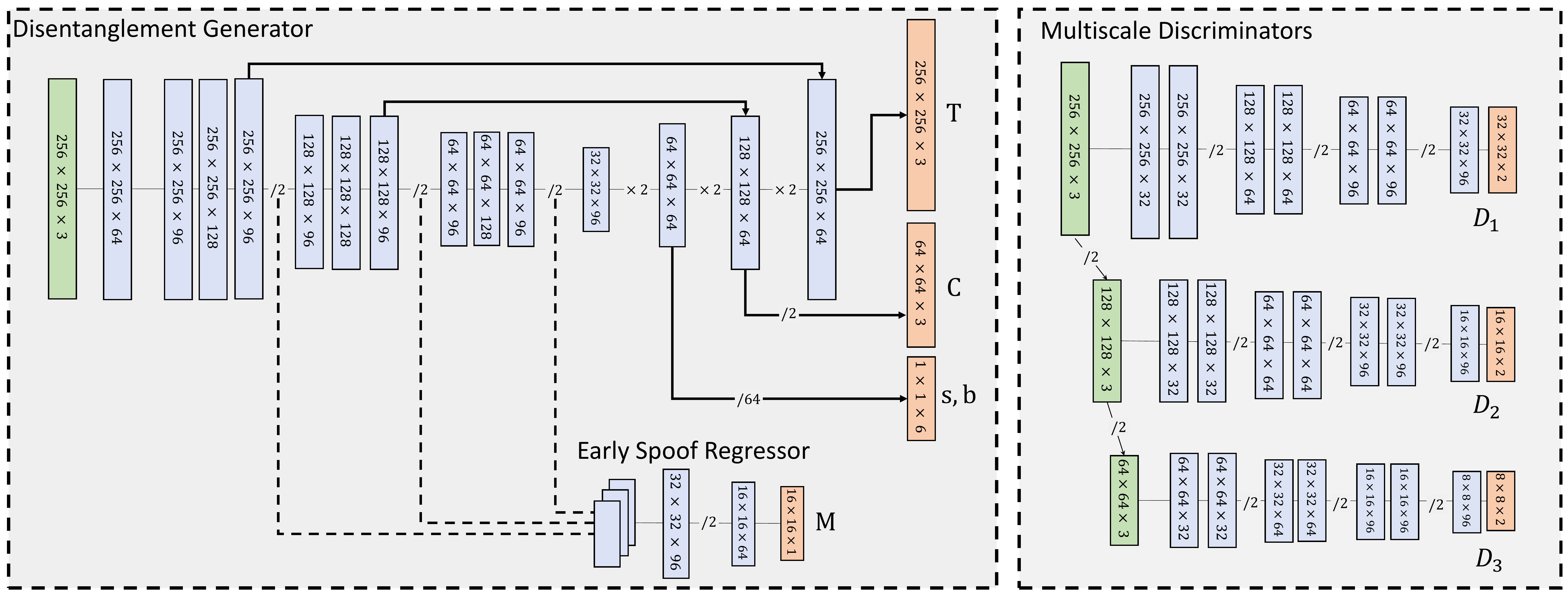}}
    \caption{\small The proposed STDN architecture. Except the last layer, each conv and transpose conv is concatenated with a Leaky ReLU layer and a batch normalization layer. $\mathbf{/2}$ denotes a downsampling by $2$, and $\mathbf{\times 2}$ denotes an upsampling by $2$.}
    \label{fig:3}
\figvspace
\end{figure*}

As shown in Fig.~\ref{fig:3}, Spoof Trace Disentanglement Network (STDN) consists of a generator and multiscale discriminators. 
They are jointly optimized to disentangle the spoof trace elements $\{\textbf{s}, \textbf{b}, \textbf{C}, \textbf{T}\}$ from the input faces.
In the rest of this section, we discuss the details of the generator, face reconstruction and synthesis, the discriminators, and the training steps and losses used in STDN.

\SubSection{Disentanglement Generator}
Spoof trace disentanglement is implemented via the generator.
The disentanglement generator adopts an encoder-decoder as the backbone network. 
The encoder progressively downsamples the input face $\mathbf{I}\! \in \!\mathbb{R}^{256\!\times \!256\!\times \!3}$ to a latent feature tensor $\mathbf{F}\! \in \!\mathbb{R}^{32\!\times \!32\!\times \!96}$ via conv layers. 
The decoder upsamples the feature tensor $\mathbf{F}$ with transpose conv layers back to the input face size.
To properly disentangle each spoof trace element, we leverage the natural upscaling property of the decoder structure:
$\textbf{s}, \textbf{b}$ have the lowest spatial resolution and thus are disentangled in the very beginning of the decoder;
$\textbf{C}$ is extracted in the middle of the decoder with the size of $64$;
$\textbf{T}$ is accordingly estimated in the last layer of the decoder.
Similar to U-Net~\cite{ronneberger2015u}, we apply the short-cut connection between encoder and decoder to leak the high-frequency details for a high-quality estimation.

Unlike typical GAN scenarios where the generator only takes data from the source domain, our generator takes data from both source (spoof) and target (live) domains, and requires high accuracy in distinguishing two domains.  
Although the spoof traces should be significantly different between the two domains, they solely are not perfect hint for classification as the intensity of spoof traces varies from type to type.
For this objective, we additionally introduce an Early Spoof Regressor (ESR) to enhance discriminativeness of the generator. 
ESR takes the bottleneck features $\mathbf{F}$ and outputs a $\mathbf{0}/\mathbf{1}$ map $\mathbf{M}\!\in\! \mathbb{R}^{16\!\times \!16}$, where $\mathbf{0}$ means live and $\mathbf{1}$ means spoof. 
Moreover, we purposely make the encoder much heavier than the decoder, \textit{i.e.}, more channels and deeper layers. This benefits the classification since ESR can better leverage the features learnt for spoof trace disentanglement.

In the testing phase, we use the average of the output from ESR and the intensity of spoof traces for classification:
\eqnvspace
\begin{equation}
\label{eq:score}
\text{score} = \frac{1}{2K^2}\|\mathbf{M}\|_1 + 
			   \frac{\alpha_0}{2N^2}\|G(\textbf{I})\|_1,
\end{equation}
where $\alpha_0$ is the weight for the spoof trace, $K\!=\!16$ is the size of $\mathbf{M}$, and $N\!=\!256$ is the image size.
\SubSection{Reconstruction and Synthesis}
There are two ways we can benefit from the spoof traces: 
\vspace{-1mm}
\begin{itemize}
\item[$\bullet$] \textbf{Reconstruction}: obtaining the live face counterpart from the input as $\hat{\textbf{I}}=\textbf{I}-G(\textbf{I})$;
\item[$\bullet$] \textbf{Synthesis}: obtaining a new spoof face by applying the spoof traces $G(\textbf{I}_i)$ disentangled from face image $\textbf{I}_i$ to a live face $\textbf{I}_j$.
\end{itemize}

Unlike the original spoof samples, the synthesized spoof come with the ground truth traces, enabling a {\it supervised} training for the generator.
However, spoof traces may contain shape-dependent content associated with the original spoof face.
Directly combining them with a live face with different shape or pose may result in poor alignment and strong visual implausibility. 
Therefore, the spoof trace should go through a geometric correction before performing the synthesis. 
We propose an online $3$D warping layer to correct the shape discrepancy.

\begin{figure}[t!]
    \centering
    \resizebox{0.85\linewidth}{!}{\includegraphics{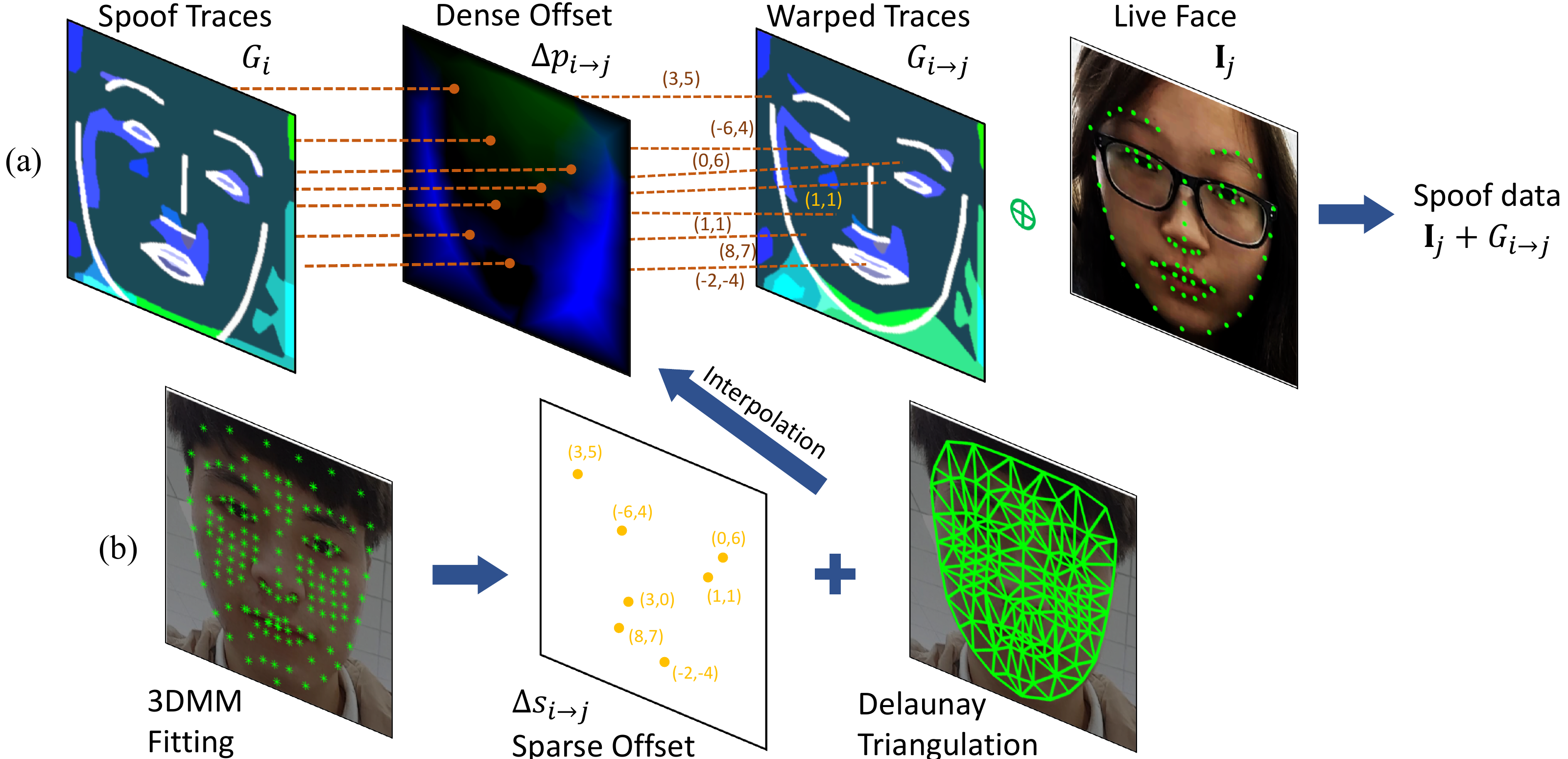}}
    \caption{\small $3$D warping pipeline. (a) Given the corresponding dense offset, we warp the spoof trace and add them to the target live face to create a new spoof. E.g. pixel $(x,y)$ with offset $(3,5)$ is warped to pixel$(x+3,y+5)$ in the new image. (b) To obtain a dense offsets from the spare offsets of the selected face shape vertices, Delaunay triangulation interpolation is adopted. }
    \label{fig:4}
\figvspace
\end{figure}

\Paragraph{Online $3$D Warping Layer}
First, the spoof traces for face $i$ can be expressed as:
\eqnvspace
\begin{equation}
    G_i = G(\textbf{I}_i)[\textbf{p}_0],
    \label{eq:g0}
\end{equation}
where $\textbf{p}_0=\{(0,0),(0,1),...,(255,255)\} \in \mathbb{R}^{256\times256\times2}$ enumerates pixel locations in $\textbf{I}_i$.
To warp the spoof trace, a dense offset $\Delta\textbf{p}_{i\rightarrow j}\in \mathbb{R}^{256\times256\times2}$ is required to indicate the offset value from face $i$ to face $j$. 
The warped traces can be denoted as:
\eqnvspace
\begin{equation}
    G_{i\rightarrow j} = G(\textbf{I}_i)[\textbf{p}_0+\Delta\textbf{p}_{i\rightarrow j}],
    \label{eq:gx2y}
\end{equation}
Since the offset $\Delta\textbf{p}_{i\rightarrow j}$  is typically composed of fractional numbers,  we implement the bilinear interpolation to sample the fractional pixel locations.
To obtain the offset $\Delta\textbf{p}_{i\rightarrow j}$, previous methods in~\cite{chang-asymmetric-style-transfer,liu-auxiliary-fas} use offline face swapping and pre-computed dense offset respectively, where both of them are non-differentiable as well as memory intensive. 
In contrast, our warping layer is both differentiable and computationally efficient, which is necessary for using it in training.
During the data preparation, we use \cite{dense-face-alignment} to fit a $3$DMM model and extract the $2$D locations of $Q$ selected vertices for each face:
\eqnvspace
\begin{equation}
    \textbf{s}=\{(x_0,y_0),(x_1,y_1),...,(x_N,y_N)\} \in \mathbb{R}^{Q\times2} ,
    \label{eq:s}
\end{equation}
A sparse offset on the corresponding vertices can then be computed between face $i$ and $j$ as $\Delta\textbf{s}_{i\rightarrow j} = \textbf{s}_j - \textbf{s}_i$.
We select $Q=140$ vertices to cover the face region so that they can represent non-rigid deformation, due to pose and expression.
To convert the sparse offset $\Delta\textbf{s}_{i\rightarrow j}$ to the dense offset $\Delta\textbf{p}_{i\rightarrow j}$, we apply a triangulation interpolation:
\eqnvspace
\begin{equation}
    \Delta\textbf{p}_{i\rightarrow j}=\text{Tri}(\textbf{p}_0, \textbf{s}_i,\Delta\textbf{s}_{i\rightarrow j}),
    \label{eq:tri}
\end{equation}
where $\text{Tri}(\cdot)$ is the interpolation operation based on Delaunay triangulation, 
Since the pixel values in the warped face are a linear combination of pixel values of the triangulation vertices, this whole process is differentiable. This process is illustrated in Fig.~\ref{fig:4}.

\Paragraph{Creating ``harder" samples}
As mentioned above, the synthesized spoof can be leveraged to enable a supervised learning for the generator.
Another advantage of the disentangled representation $\{\textbf{s},\textbf{b},\textbf{C},\textbf{T}\}$ is that we can manipulate the spoof traces via tuning these elements, such as diminishing or amplifying any certain element.
While diminishing one or a few elements in $\{\textbf{s},\textbf{b},\textbf{C},\textbf{T}\}$, the synthesized spoof becomes ``less spoofed", and thus closer to a live face since the spoof traces are weakened.
Such spoof data can be regarded as \textit{harder} samples and may benefit the learning of the generator.
\textit{E.g.}, while removing the color distortion \textbf{s} from a replay spoof trace, the generator may be forced to rely on other elements such as high-level texture patterns.
In this work, we randomly set one element from $\{\textbf{s},\textbf{b},\textbf{C},\textbf{T}\}$ to be zero when synthesizing a new spoof face.
Compared with other methods, such as brightness and contrast change~\cite{presentation-attack-detection-for-face-in-mobile-phones}, reflection and blurriness effect~\cite{yang2019face}, or $3$D distortion~\cite{guo2019improving}, our approach can introduce more realistic and effective data samples, as shown in Sec.~\ref{sec:4}.

\SubSection{Multi-scale Discriminators}
Motivated by~\cite{wang2018pix2pixHD}, we adopt three discriminators $D_1$, $D_2$, and $D_3$ at different resolutions (\textit{i.e.}, $256$, $128$, and $64$) in our GAN architecture.
The faces in the original size are sent to $D_1$, resized by a ratio of $2$ and sent to $D_2$, and resized by a ratio of $4$ and sent to $D_3$.
$D_1$, working in the highest scale, focuses on the fine texture details.
$D_2$, working in the middle scale, focuses more on the content pattern in $\textbf{C}$.
$D_3$, working in the lowest scale, focuses on global elements since the higher-frequency detail in $\textbf{C}$ and $\textbf{T}$ might be erased by resizing.
For each discriminator, we adopt the structure of PatchGAN~\cite{isola2017image}, which essentially is a fully convolutional network. Fully convolutional networks are shown to be effective to not only synthesize high-quality images~\cite{isola2017image,wang2018pix2pixHD}, but also tackle face anti-spoofing problems~\cite{liu-auxiliary-fas}.
Specifically, each discriminator consists of $7$ conv and $3$ downsampling layers.
It outputs a $2$-channel map, where each channel represents output of one domain (\textit{i.e.}, live and spoof).
The first channel compares the reconstructed live samples with the real live samples, while the second channel compares the synthesized spoof samples with real spoof samples.
\SubSection{Training Steps and Loss Functions}
\label{sec:3-loss}
We utilize multiple loss functions in our three training steps. We will introduce them first, followed by how they are used in the training steps.

\noindent\textbf{ESR loss:} 
For live faces, $\mathbf{M}$ should be zero, and for spoof faces as well as synthesized spoof faces, $\mathbf{M}$ should be one. We apply the $\mathcal{L}$-$1$ norm on this loss as:  
\eqnvspace
\begin{equation}
    L_{\textit{ESR}} = \frac{1}{K^2}(\mathbb{E}_{i\sim \mathcal{L}}
    [{\lVert}\mathbf{M}_i{\rVert}_1] + \mathbb{E}_{i\sim \mathcal{S}\cup\hat{\mathcal{S}}}
    [{\lVert}\mathbf{M}_i-1{\rVert}_1]),
    \label{eq:ploss}
\end{equation}
where $\hat{\mathcal{S}}$ denotes the domain of synthesized spoof faces and $K\!=\!16$ is the size of $\mathbf{M}$.

\noindent\textbf{Adversarial loss for $G$:} 
We employ the LSGANs~\cite{mao2017least} on reconstructed live and synthesized spoof. It pushes the reconstructed live faces to domain $\mathcal{L}$, and the synthesized spoof faces to domain $\mathcal{S}$:
\eqnvspace
\begin{equation}
    \begin{aligned}
    L_{G} = \sum_{n=1,2,3} \{\mathbb{E}_{i\sim \mathcal{S}}
        [ (D^1_n(\textbf{I}_i\!-\!G_i)\!- \!\mathbf{1})^2]
        +\mathbb{E}_{i\sim \mathcal{L}, j\sim \mathcal{S}}
        [(D^2_n(\textbf{I}_i\!+\!G_{j\rightarrow i})\! -\mathbf{1})^2]\},
    \end{aligned}
    \label{eq:lg}
\end{equation}
where $D^1_n$ and $D^2_n$ denote the first and second channel of discriminator $D_n$.

\noindent\textbf{Adversarial loss for $D$:} 
The adversarial loss pushes the discriminators to distinguish between real live \textit{vs.}~reconstructed live, and real spoof \textit{vs.}~synthesized spoof:
\eqnvspace
\begin{equation}\label{eq:ld}
\begin{aligned}
L_{D}\! =\! \sum_{n=1,2,3}\! \{\mathbb{E}_{i\sim \mathcal{L}}
    [(D^1_n(\textbf{I}_i) \!-\! \mathbf{1})^2]\! +\!
    \mathbb{E}_{i\sim \mathcal{S}}
    [ (D^2_n(\textbf{I}_i)\!-\!\mathbf{1})^2]     \\
    +\mathbb{E}_{i\sim \mathcal{S}}
    [ (D^1_n(\textbf{I}_i\!-\!G_i(x)))^2] +
    \mathbb{E}_{i\sim \mathcal{L}, j\sim \mathcal{S}}
    [D^2_n(\textbf{I}_i\!+\!G_{j\rightarrow i})\!)^2]\}.
\end{aligned}
\end{equation}

\noindent\textbf{Regularizer loss:} 
In Eq.~\ref{eq:op}, the task regularizes the intensity of spoof traces  while satisfying certain domain conditions.
This regularizer loss is denoted as: 
\eqnvspace
\begin{equation}
    L_{R} = \beta\,\mathbb{E}_{\textbf{x}\sim \mathcal{L}}
    [{\lVert}G(\textbf{I}_i){\rVert}_2^2] +
    \mathbb{E}_{\textbf{i}\sim \mathcal{S}}
    [{\lVert}G(\textbf{I}_i){\rVert}_2^2],
    \label{eq:ploss}
\end{equation}
where $\beta>1$ is a weight to further compress the traces of live faces to be zero.

\noindent\textbf{Pixel loss:} 
Synthesized spoof data come with ground truth spoof traces. Therefore we can enable a supervised pixel loss for the generator to disentangle the exact spoof traces that were added to the live faces:
\eqnvspace
\begin{equation}
    L_{P} = \mathbb{E}_{\textbf{i}\sim \mathcal{L}, \textbf{j}\sim \mathcal{S}}
    [{\lVert}G(\lceil \textbf{I}_i + G_{j\rightarrow i}  \rceil) - 
    \lceil G_{j\rightarrow i}\rceil {\rVert}_1],
    \label{eq:ploss}
\end{equation}
where $\lceil \cdot \rceil$ is the \verb|stop_gradient| operation. In this loss, we regard the traces $G_{j\rightarrow i}$ as ground truth, and the \verb|stop_gradient| operation can prevent changing $G_{j\rightarrow i}$ to minimize the loss.

\begin{figure*}[t!]
    \centering
    \resizebox{1\linewidth}{!}{\includegraphics{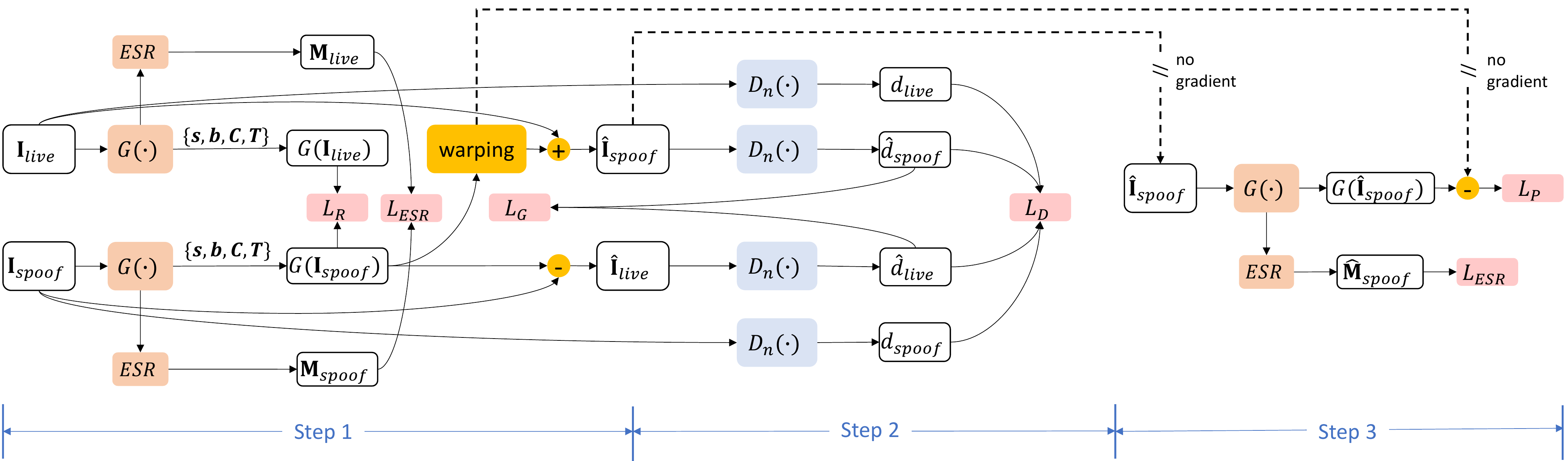}}
    \caption{\small The three training steps of STDN. Each mini-batch includes the same number of live and spoof samples.}
    \label{fig:step}
    \vspace{-2mm}
\end{figure*}

\noindent\textbf{Training steps and total loss:} 
Shown in Fig.~\ref{fig:step}, each mini-batch has $3$ training steps: generator step, discriminator step, and extra supervision step.
In the generator step, live faces $\textbf{I}_{\textit{live}}$ and spoof faces $\textbf{I}_{\textit{spoof}}$ are fed to generator $G(\cdot)$ to disentangle the spoof traces. The spoof traces are used to reconstruct the live counterpart $\hat{\textbf{I}}_{\textit{live}}$ and synthesize new spoof $\hat{\textbf{I}}_{\textit{spoof}}$. The generator is updated with respect to adversarial loss $L_G$, ESR loss $L_{\textit{ESR}}$, and regularizer loss $L_R$:
\eqnvspace
\begin{equation}
    L =  \alpha_1L_G + \alpha_2 L_{\textit{ESR}} + \alpha_3 L_R.
    \label{eq:loss_overall}
\end{equation}
For the discriminator step, $\textbf{I}_{\textit{live}}$, $\textbf{I}_{\textit{spoof}}$, $\hat{\textbf{I}}_{\textit{live}}$, and $\hat{\textbf{I}}_{\textit{spoof}}$ are fed into the discriminators $D_n(\cdot), n\!=\!\{1,2,3\}$. The discriminators are supervised with adversarial loss $L_D$ to compete with the generator.
For the extra supervision step, $\textbf{I}_{\textit{live}}$ and $\hat{\textbf{I}}_{\textit{spoof}}$ are fed into the generator with ground truth label and trace to enable pixel loss $L_P$ and ESR loss $L_{\textit{ESR}}$:
\eqnvspace
\begin{equation}
    L = \alpha_4 L_{\textit{ESR}} + \alpha_5L_P,
    \label{eq:loss_overall}
\end{equation}
where $\alpha_1$-$\alpha_5$ are the weights to balance the multitask training. 
To note that, in the extra supervision step, we send the original live faces $\textbf{I}_{\textit{live}}$ with $\hat{\textbf{I}}_{\textit{spoof}}$ for a balanced mini-batch, which is important when computing the moving average in the batch normalization layer.
We execute all $3$ steps in each minibatch iteration, but reduce the learning rate for discriminator step by half.

\Section{Experiments}
\label{sec:4}
In this section, we first introduce the experiments setup, and then present the performance in both the known spoof and unknown spoof scenarios.
Next, we quantitatively evaluate the spoof traces by performing a spoof medium classification, and conduct an ablation study on each design in the proposed method.
Finally, we provide visualization results on the spoof trace disentanglement and new spoof synthesis.
\SubSection{Experimental Setup}
\Paragraph{Databases}
We conduct experiments on three major databases: Oulu-NPU~\cite{OULU_NPU_2017}, SiW~\cite{liu-auxiliary-fas}, and SiW-M~\cite{deep-tree}.
Oulu-NPU and SiW include print/replay attacks, while SiW-M includes $13$ spoof types.
We follow all the testing protocols and compare with SOTA methods.
Similar to most prior works, we only use the face region for training and testing.

\Paragraph{Evaluation metrics}
Two standard metrics are used in this work for comparison: EER and APCER/BPCER/ACER.
EER describes the theoretical performance and predetermines the threshold for making decisions.
APCER/BPCER/ACER\cite{acer1} describe the performance given a predetermined threshold.
For EER/ACER, the lower the better.
We also report the True Detection Rate (TDR) at a given False Detection Rate (FDR). This metric describes the spoof detection rate at a strict tolerance to live errors, which is widely used to evaluate  systems in real-world applications~\cite{iarpa-odin}. In this work, we report TDR at FDR$=0.5\%$.
For TDR, the higher the better.

\Paragraph{Parameter setting}
STDN is implemented in Tensorflow with an initial learning rate of $1e$-$4$.
We train in total $150,000$ iterations with a batch size of $8$, and decrease the learning rate by a ratio of $10$ every $45,000$ iterations. 
We initialize the weights with $[0,0.02]$ normal distribution. 
$\{\alpha_1,\alpha_2,\alpha_3,\alpha_4,\alpha_5,\beta\}$ are set to be $\{1,100,1e$-$3,50,1,1e4\}$.
$\alpha_0$ is empirically determined from the training or validation set.
We use open source face alignment~\cite{bulat2017far} and $3$DMM fitting~\cite{dense-face-alignment} to crop the face and provide $140$ landmarks.

\begin{table}[t!]
\small
\centering
\caption{\small Known spoof detection on: (a) OULU-NPU (b) SiW (c) SiW-M Protocol I.}
	
\begin{tabular}{ccc}
	\scalebox{0.65}{
	\begin{tabular}{cllll}
        \toprule Protocol & Method & APCER (\%) & BPCER (\%) & ACER (\%) \\ \hline
        \multirow{4}{*}{1} 
               & STASN\cite{yang2019face} & $1.2$ & $2.5$ & $1.9$ \\
               & Auxiliary~\cite{liu-auxiliary-fas} & $1.6$ & $1.6$ & $1.6$ \\
               & DeSpoof~\cite{jourabloo-face-despoofing} & $1.2$ & $1.7$ & $1.5$ \\
               & Ours & $\textbf{0.8}$ & $\textbf{1.3}$ & $\textbf{1.1}$ \\ \hline
        \multirow{4}{*}{2} 
               & Auxiliary~\cite{liu-auxiliary-fas} & $2.7$ & $2.7$ & $2.7$ \\
               & GRADIANT~\cite{OULU_NPU_2017} & $3.1$ & $1.9$ & $2.5$ \\
               & STASN\cite{yang2019face} & $4.2$ & $\textbf{0.3}$ & $2.2$ \\
               & Ours & $\textbf{2.3}$ & $1.6$ & $\textbf{1.9}$ \\ \hline
        \multirow{4}{*}{3} 
               & DeSpoof~\cite{jourabloo-face-despoofing} & $4.0\pm1.8$ & $3.8\pm1.2$ & $3.6\pm1.6$ \\
               & Auxiliary~\cite{liu-auxiliary-fas} & $2.7\pm1.3$ & $3.1\pm1.7$ & $2.9\pm1.5$ \\
               & STASN\cite{yang2019face} & $4.7\pm3.9$ & $\textbf{0.9}\pm\textbf{1.2}$ & $\textbf{2.8}\pm\textbf{1.6}$ \\
               & Ours & $\textbf{1.6}\pm \textbf{1.6}$ & $4.0\pm 5.4$ & $\textbf{2.8}\pm \textbf{3.3}$ \\ \hline
        \multirow{4}{*}{4} 
               & Auxiliary~\cite{liu-auxiliary-fas} & $9.3\pm5.6$ & $10.4\pm6.0$ & $9.5\pm6.0$ \\
               & STASN\cite{yang2019face} & $6.7\pm10.6$ & $8.3\pm8.4$ & $7.5\pm4.7$ \\
               & DeSpoof~\cite{jourabloo-face-despoofing} & $5.1\pm6.3$ & $6.1\pm5.1$ & $5.6\pm5.7$ \\
               & Ours & $\textbf{2.3}\pm \textbf{3.6}$ & $\textbf{5.2}\pm \textbf{5.4}$ & $\textbf{3.8}\pm \textbf{4.2}$ \\ \bottomrule 
        \vspace{-2mm} \\
               \multicolumn{5}{c}{\large{(a)}}
    \end{tabular}}
    
    & \qquad \qquad& 

	\scalebox{0.65}{
	\begin{tabular}{cllll}
        \toprule Protocol & Method & APCER (\%) & BPCER (\%) & ACER (\%) \\ \hline
        \multirow{4}{*}{1} 
            & Auxiliary\cite{liu-auxiliary-fas} & $3.6$ & $3.6$ & $3.6$ \\
            & STASN\cite{yang2019face} & $-$ & $-$ & $1.0$ \\
            & Meta-FAS-DR\cite{zhao2019meta} & $0.5$ & $0.5$ & $0.5$ \\
            & Ours & $\textbf{0.0}$ & $\textbf{0.0}$ & $\textbf{0.0}$ \\ \hline
        \multirow{4}{*}{2} 
            & Auxiliary\cite{liu-auxiliary-fas} & $0.6\pm0.7$ & $0.6\pm0.7$ & $0.6\pm0.7$ \\
            & Meta-FAS-DR\cite{zhao2019meta} & $0.3\pm0.3$ & $0.3\pm0.3$ & $0.3\pm0.3$ \\
            & STASN\cite{yang2019face} & $-$ & $-$ & $0.3\pm0.1$ \\
            & Ours & $\textbf{0.0}\pm\textbf{0.0}$ & $\textbf{0.0}\pm\textbf{0.0}$ & $\textbf{0.0}\pm\textbf{0.0}$ \\ \hline
        \multirow{4}{*}{3} 
            & STASN\cite{yang2019face} & $-$ & $-$ & $12.1\pm1.5$ \\
            & Auxiliary\cite{liu-auxiliary-fas} & $8.3\pm3.8$ & $8.3\pm3.8$ & $8.3\pm3.8$ \\
            & Meta-FAS-DR\cite{zhao2019meta} & $\textbf{8.0}\pm\textbf{5.0}$ & $\textbf{7.4}\pm\textbf{5.7}$ & $\textbf{7.7}\pm\textbf{5.3}$ \\
            & Ours & $8.3\pm 3.3$ & $7.5\pm 3.3$ & $7.9\pm 3.3$ \\ \bottomrule
        \vspace{-2mm} \\
            \multicolumn{5}{c}{\large{(b)}}
    \end{tabular}}\\
    
    \multicolumn{3}{c}{\scalebox{0.7}{
	\begin{tabular}{llllllllllllllllll}
	\multicolumn{17}{c}{} \\ \toprule
	\multirow{2}{*}{Metrics(\%)} &\multirow{2}{*}{Replay}& \multirow{2}{*}{Print} & \multicolumn{5}{c}{3D Mask}  && \multicolumn{3}{c}{Makeup}  &&  \multicolumn{3}{c}{Partial Attacks} & \multirow{2}{*}{Overall}\\ 
	\cline{4-8} \cline{10-12} \cline{14-16}
	& & & Half & Silic. & Trans. & Paper & Manne. && Ob. & Im. & Cos. && Funny. & Papergls. & Paper &\\ \hline
	\multicolumn{17}{c}{ACER(\%)} \\ \hline
    Auxiliary\cite{liu-auxiliary-fas}
        &$5.1$& $5.0$ & $5.0$ & $10.2$ & $5.0$ & $9.8$ & $6.3$ && $19.6$ & $5.0$ & $26.5$ && $5.5$ & $5.2$ & $5.0$ & $6.3$ \\
    Ours
        &$\textbf{3.2}$& $\textbf{3.1}$ & $\textbf{3.0}$ & $\textbf{9.0}$ & $\textbf{3.0}$ & $\textbf{3.4}$ & $\textbf{4.7}$ && $\textbf{3.0}$ & $\textbf{3.0}$ & $\textbf{24.5}$ && $\textbf{4.1}$ & $\textbf{3.7}$ & $\textbf{3.0}$ & $\textbf{4.1}$ \\ \hline
	\multicolumn{17}{c}{EER(\%)} \\ \hline
    Auxiliary\cite{liu-auxiliary-fas}
        &$4.7$& $0.0$ & $1.6$ & $10.5$ & $4.6$ & $10.0$ & $6.4$ && $12.7$ & $0.0$ & $19.6$ && $7.2$ & $7.5$ & $0.0$ & $6.6$ \\
    Ours
        &$\textbf{2.1}$& $\textbf{2.2}$ & $\textbf{0.0}$ & $\textbf{7.2}$ & $\textbf{0.1}$ & $\textbf{3.9}$ & $\textbf{4.8}$ && $\textbf{0.0}$ & $\textbf{0.0}$ & $\textbf{19.6}$ && $\textbf{5.3}$ & $\textbf{5.4}$ & $\textbf{0.0}$ & $\textbf{4.8}$ \\ \hline
    \multicolumn{17}{c}{TDR@FDR=$0.5$(\%)} \\ \hline
    Ours
        &$90.1$& $76.1$ & $80.7$ & $71.5$ & $62.3$ & $74.4$ & $85.0$ && $100.0$ & $100.0$ & $33.8$ && $49.6$ & $30.6$ & $97.7$ & $70.4$ \\ \bottomrule
        \vspace{-2mm} \\
    \multicolumn{17}{c}{\large{(c)}}
	\end{tabular}}}
\label{tab:known}
\end{tabular}
\figvspace
\end{table}
\SubSection{Anti-Spoofing for Known Spoof Types}
\Paragraph{Oulu-NPU~\cite{OULU_NPU_2017}}
is a commonly used face anti-spoofing benchmark due to its high quality and challenging testing.
Shown in Tab.~\ref{tab:known}(a), our approach achieves the best performance in all four protocols. Specifically, we demonstrate significant improvement in protocol $1$ and protocol $4$, reducing the ACER by $30\%$ and $32\%$ relative to the best prior work.
However, in protocol $3$ and protocol $4$, the performances of testing camera $6$ are much lower than those of cameras $1$-$5$: the ACER for camera $6$ are $9.5\%$ and $8.6\%$, while the average ACER for the other cameras are $1.7\%$ and $3.1\%$ respectively. 
Compared with other cameras, we notice that camera $6$ has stronger sensor noises and STDN recognizes them as unknown spoof traces, which leads to an increasing BPCER.
Separating sensor noises from spoof traces can be an important future research topic. 
\begin{table*}[t!]
\small
\centering
\caption{\small The evaluation on SiW-M Protocol II: unknown spoof detection. \textbf{Bold} indicates the best score in each protocol. \textbf{\textcolor{red}{Red}} indicates protocols that our method improves over $50\%$ than SOTA.}
\resizebox{1\textwidth}{!}{
	\begin{tabular}{lllllllllllllllll}
	\multicolumn{15}{c}{} \\ \toprule
	\multirow{2}{*}{Methods} &\multirow{2}{*}{Replay}& \multirow{2}{*}{Print} & \multicolumn{5}{c}{3D Mask}  && \multicolumn{3}{c}{Makeup}  &&  \multicolumn{3}{c}{Partial Attacks} & \multirow{2}{*}{Average}\\
	\cline{4-8} \cline{10-12} \cline{14-16}
	& & & Half & Silic. & Trans. & Paper & Manne. && Ob. & Im. & Cos. && Fun. & Papergls. & Paper &\\ \hline
	\multicolumn{17}{c}{APCER(\%)} \\ \hline
    LBP+SVM~\cite{OULU_NPU_2017}
        &$19.1$& $15.4$ & $40.8$ & $20.3$ & $70.3$ & $\mathbf{0.0}$ & $4.6$ && $96.9$ & $35.3$ & $\mathbf{11.3}$ && $53.3$ & $58.5$ & $0.6$ & $32.8\pm29.8 $ \\ 
    Auxiliary\cite{liu-auxiliary-fas}
        &$23.7$& $7.3$ & $27.7$ & $18.2$ & $97.8$ & $8.3$ & $16.2$ && $100.0$ & $18.0$ & $16.3$ && $91.8$ & $72.2$ & $0.4$ & $38.3\pm37.4$  \\ 
    DTL \cite{deep-tree}
        &$\textbf{1.0}$& $\mathbf{0.0}$ & $0.7$ & $24.5$ & $58.6$ & $0.5$ & $3.8$ && $\mathbf{73.2}$ & $13.2$ & $12.4$ && $17.0$ & $17.0$ & $0.2$ & $17.1\pm23.3$  \\ 
    Ours
        &$1.6$& $\mathbf{0.0}$ & $\mathbf{0.5}$ & $\mathbf{\textcolor{red}{7.2}}$ & $\mathbf{\textcolor{red}{9.7}}$ & $0.5$ & $\mathbf{\textcolor{red}{0.0}}$ && $96.1$ & $\mathbf{\textcolor{red}{0.0}}$ & $21.8$ && $\mathbf{14.4}$ & $\mathbf{\textcolor{red}{6.5}}$ & $\mathbf{0.0}$ & $\mathbf{12.2}\pm\mathbf{26.1}$  \\ \hline
	\multicolumn{17}{c}{BPCER(\%)} \\ \hline
    LBP+SVM~\cite{OULU_NPU_2017}
     &$22.1$& $21.5$ & $21.9$ & $21.4$ & $20.7$ & $23.1$ & $22.9$ && $21.7$ & $12.5$ & $22.2$ && $18.4$ & $20.0$ & $22.9$ & $21.0\pm2.9$  \\ 
    Auxiliary\cite{liu-auxiliary-fas} 
        &$\mathbf{10.1}$& $\mathbf{6.5}$ & $\mathbf{10.9}$ & $\mathbf{11.6}$ & $\mathbf{6.2}$ & $\mathbf{7.8}$ & $\mathbf{9.3}$ && $11.6$ & $9.3$ & $\mathbf{7.1}$ && $\mathbf{6.2}$ & $\mathbf{8.8}$ & $\mathbf{10.3}$ & $\mathbf{8.9}\pm\mathbf{2.0}$  \\ 
    DTL \cite{deep-tree}
        &$18.6$& $11.9$ & $29.3$ & $12.8$ & $13.4$ & $8.5$ & $23.0$ && $11.5$ & $9.6$ & $16.0$ && $21.5$ & $22.6$ & $16.8$ & $16.6\pm6.2$ \\ 
    Ours
        &$14.0$& $14.6$ & $13.6$ & $18.6$ & $18.1$ & $8.1$ & $13.4$ && $\mathbf{10.3}$ & $\mathbf{9.2}$ & $17.2$ && $27.0$ & $35.5$ & $11.2$ & $16.2\pm7.6$ \\ \hline
	\multicolumn{17}{c}{ACER(\%)} \\ \hline
    LBP+SVM~\cite{OULU_NPU_2017}
        &$20.6$& $18.4$ & $31.3$ & $21.4$ & $45.5$ & $11.6$ & $13.8$ && $59.3$ & $23.9$ & $16.7$ && $35.9$ & $39.2$ & $11.7$ & $26.9\pm14.5$  \\ 
    Auxiliary\cite{liu-auxiliary-fas}
        &$16.8$& $6.9$ & $19.3$ & $14.9$ & $52.1$ & $8.0$ & $12.8$ && $55.8$ & $13.7$ & $\mathbf{11.7}$ && $49.0$ & $40.5$ & $\mathbf{5.3}$ &$23.6\pm18.5$ \\ 
    DTL \cite{deep-tree}
        &$9.8$& $\mathbf{6.0}$ & $15.0$ & $18.7$ & $36.0$ & $4.5$ & $13.4$ && $\mathbf{48.1}$ & $11.4$ & $14.2$ && $\mathbf{19.3}$ & $\mathbf{19.8}$ & $8.5$ & $16.8\pm11.1$  \\ 
    Ours
        &$\mathbf{7.8}$& $7.3$ & $\mathbf{\textcolor{red}{7.1}}$ & $\mathbf{12.9}$ & $\mathbf{\textcolor{red}{13.9}}$ & $\mathbf{4.3}$ & $\mathbf{\textcolor{red}{6.7}}$ && $53.2$ & $\mathbf{\textcolor{red}{4.6}}$ & $19.5$ && $20.7$ & $21.0$ & $5.6$ & $\mathbf{14.2}\pm\mathbf{13.2}$  \\ \hline
	\multicolumn{17}{c}{EER(\%)} \\ \hline
    LBP+SVM~\cite{OULU_NPU_2017}
        &$20.8$& $18.6$ & $36.3$ & $21.4$ & $37.2$ & $7.5$ & $14.1$ && $51.2$ & $19.8$ & $16.1$ && $34.4$ & $33.0$ & $7.9$ & $24.5\pm12.9$ \\ 
    Auxiliary\cite{liu-auxiliary-fas}
        &$14.0$& $4.3$ & $11.6$ & $\mathbf{12.4}$ & $24.6$ & $7.8$ & $10.0$ && $72.3$ & $10.1$ & $\mathbf{9.4}$ && $21.4$ & $\mathbf{18.6}$ & $4.0$  & $17.0\pm17.7$ \\ 
    DTL \cite{deep-tree}
        &$10.0$& $\mathbf{2.1}$ & $14.4$ & $18.6$ & $26.5$ & $5.7$ & $9.6$ && $50.2$ & $10.1$ & $13.2$ && $\mathbf{19.8}$ & $20.5$ & $8.8$ & $16.1\pm12.2$ \\ 
    Ours
        &$\mathbf{7.6}$& $3.8$ & $\mathbf{8.4}$ & $13.8$ & $\mathbf{\textcolor{red}{14.5}}$ & $\mathbf{5.3}$ & $\mathbf{\textcolor{red}{4.4}}$ && $\mathbf{35.4}$ & $\mathbf{\textcolor{red}{0.0}}$ & $19.3$ && $21.0$ & $20.8$ & $\mathbf{\textcolor{red}{1.6}}$ & $\mathbf{12.0}\pm\mathbf{10.0}$ \\ \hline
    \multicolumn{17}{c}{TDR@FDR=$0.5$(\%)} \\ \hline
    Ours
        &$45.0$& $40.5$ & $45.7$ & $36.7$ & $11.7$ & $40.9$ & $74.0$ && $0.0$ & $67.5$ & $16.0$ && $13.4$ & $9.4$ & $62.8$ & $35.7\pm23.9$ \\ \bottomrule
	\end{tabular}
\label{tab:siwm_p2}} 
\figvspace
\end{table*}

\begin{figure*}[t]
\centering
\resizebox{1\linewidth}{!}{\includegraphics{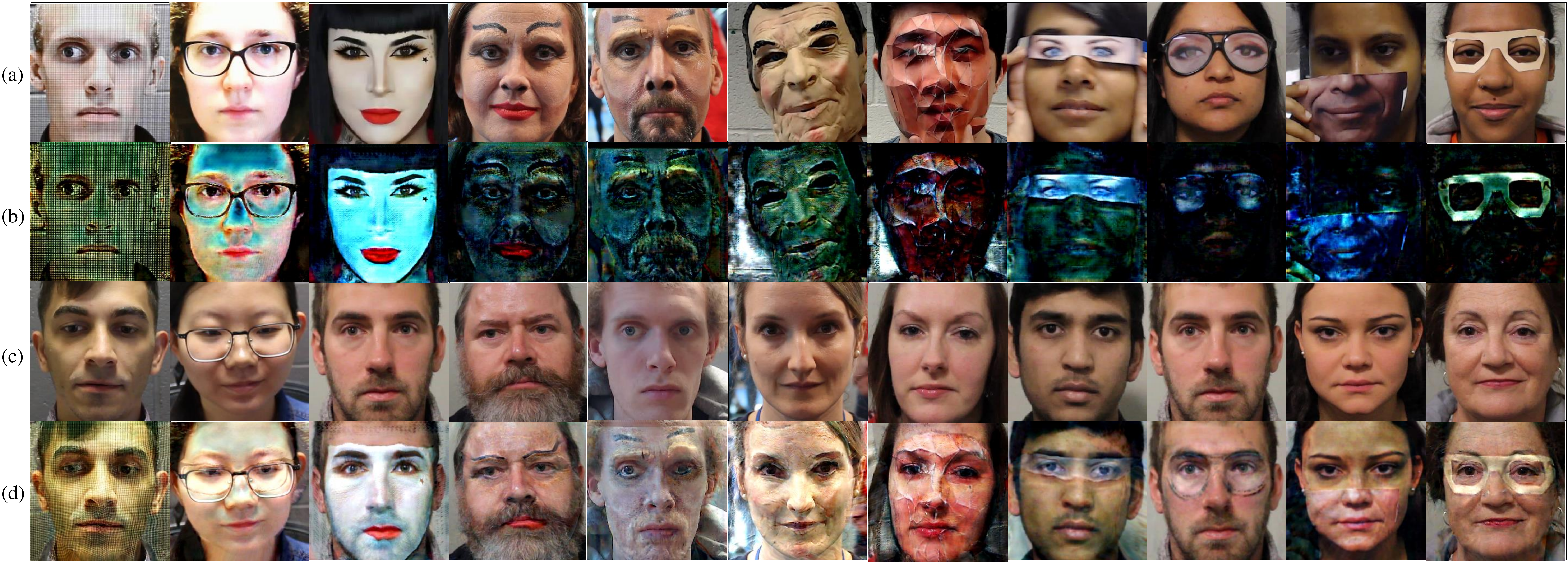}}
\caption{\small Examples of the spoof data synthesis. (a) The source spoof samples $\textbf{I}_i$. (b) The disentangled spoof traces $G(\textbf{I}_i)$. (c) The target live faces $\textbf{I}_j$. (d) The synthesized spoof $\textbf{I}_j+G_{i\rightarrow j}$.}
\label{fig:7}
\figvspace
\end{figure*}

\Paragraph{SiW~\cite{liu-auxiliary-fas}}
is another recent high-quality database.
It includes fewer capture cameras but more spoof mediums and environment variations, such as pose, illumination, and expression.
The comparison on three protocols is shown in Tab.~\ref{tab:known}(b).
We outperform the previous works on the first two protocols and have a competitive performance on protocol $3$. 
Protocol $3$ aims to test the performance of unknown spoof detection, where the model is trained on one spoof attack (print or replay) and tested on the other.
As we can see from Fig.~\ref{fig:7}, the traces of print and replay are significantly different, which would prevent the model from generalizing well.

\Paragraph{SiW-M~\cite{deep-tree}}
contains a large diversity of spoof types, including print, replay, $3$D mask, makeup, and partial attacks.
This allows us to have a comprehensive evaluation of the proposed approach with different spoof attacks.
To use SiW-M, we randomly split the data into train/test set with a ratio of $60\%$ and $40\%$, and the results are shown in Tab.~\ref{tab:known}(c).
Compared to one of the best anti-spoofing models~\cite{liu-auxiliary-fas}, our method outperforms on all spoof types as well as the overall performance, which demonstrates the superiority of our anti-spoofing  on known spoof attacks.

\SubSection{Anti-Spoofing for Unknown Spoof Types}
Another important aspect of anti-spoofing model is to generalize to the unknown/unseen.
SiW-M comes with the testing protocol to evaluate the performance of unknown attack detection.
Shown in Tab.~\ref{tab:siwm_p2}, STDN achieves significant improvement over the previous best model by relatively $24.8\%$ on the overall EER and $15.5\%$ on the overall ACER.
This is especially noteworthy because DTL was specifically designed for detecting unknown spoof types, while our proposed approach shines in {\it both known and unknown spoof detection}.
Specifically, we reduce the EERs of transparent mask, mannequin head, impersonation makeup and partial paper attack relatively by $45.3\%$, $54.2\%$, $100.0\%$, $81.8\%$, respectively.
Among all, obfuscation makeup is the most challenging one, where we predict almost all the spoof samples as live. This is due to the fact that such makeup looks very similar to the live faces, while being dissimilar to any other spoof types. Once we obtain a few samples, our model can quickly recognize the spoof traces on the eyebrow and cheek, and successfully detect the attack ($0 \%$ in Tab.~\ref{tab:known}(c)).
However, with the TDR$=35.7\%$ at FDR$=0.5\%$, the proposed method is still far from applicable in practices when dealing with unknown spoof types, which warrant future research. 
\begin{table}[t!]\small\centering
\caption{\small Confusion matrices of spoof mediums classification based on spoof traces. The left table is $3$-class classification, and the right is $5$-class classification. The results are compared with the previous method~\cite{jourabloo-face-despoofing}. \textcolor{ForestGreen}{Green} represents improvement over~\cite{jourabloo-face-despoofing}.  \textcolor{red}{Red} represents performance drop.}
\begin{tabular}{ccc}
	\scalebox{0.75}{\begin{tabular}{llll}
		\toprule
		\backslashbox{Label}{Predict} & Live & Print & Replay\\ \midrule	
		Live & \cellcolor{blue!12}$60(\textcolor{ForestGreen}{+1})$ & $0(\textcolor{ForestGreen}{-1})$  & $0$ \\ 
		Print & $3(\textcolor{red}{+3})$ & \cellcolor{blue!12}$108(\textcolor{ForestGreen}{+20})$ & $9(\textcolor{ForestGreen}{-23})$ \\ 
		Replay & $1(\textcolor{ForestGreen}{-12})$ & $11(\textcolor{red}{+3})$  & \cellcolor{blue!12}$108(\textcolor{ForestGreen}{+9})$ \\ \bottomrule
    \end{tabular}}

& \qquad \qquad \qquad & 

	\scalebox{0.6}{\begin{tabular}{llllll}
		\toprule
		\backslashbox{Label}{Predict}& Live & Print$1$ & Print$2$ & Replay$1$ & Replay$2$\\ \midrule
		Live       & \cellcolor{blue!12}$56(\color{red}{-4}$$)$ & $1(\color{red}{+1}$$)$  & $1(\color{red}{+1}$$)$  & $1(\color{red}{+1}$$)$  & $1(\color{red}{+1}$$)$ \\ 
		Print$1$   & $0$  & \cellcolor{blue!12}$43(\textcolor{ForestGreen}{+2})$ & $11(\textcolor{red}{+9})$  & $3(\textcolor{ForestGreen}{-8})$ & $3(\textcolor{ForestGreen}{-3})$ \\ 
		Print$2$   & $0$  & $9(\textcolor{ForestGreen}{-25})$ & \cellcolor{blue!12}$48(\textcolor{ForestGreen}{+37})$ & $1(\textcolor{ForestGreen}{-8})$  & $2(\textcolor{ForestGreen}{-4})$ \\ 
		Replay$1$ & $1(\textcolor{ForestGreen}{-9})$ & $2(\textcolor{ForestGreen}{-1})$  & $3(\textcolor{red}{+3})$  & \cellcolor{blue!12}$51(\textcolor{ForestGreen}{+38})$ & $3(\textcolor{ForestGreen}{-28})$ \\ 
		Replay$2$ & $1(\textcolor{ForestGreen}{-7})$  & $2(\textcolor{ForestGreen}{-5})$  & $2(\textcolor{red}{+2})$  & $3(\textcolor{ForestGreen}{-3})$  & \cellcolor{blue!12}$52(\textcolor{ForestGreen}{+13})$ \\ \bottomrule
	\end{tabular}
\label{tab:cls}}
	\end{tabular}
\figvspace
\end{table}

\begin{figure*}[t]
\begin{floatrow}
\ffigbox{%
  \resizebox{\linewidth}{!}{\includegraphics{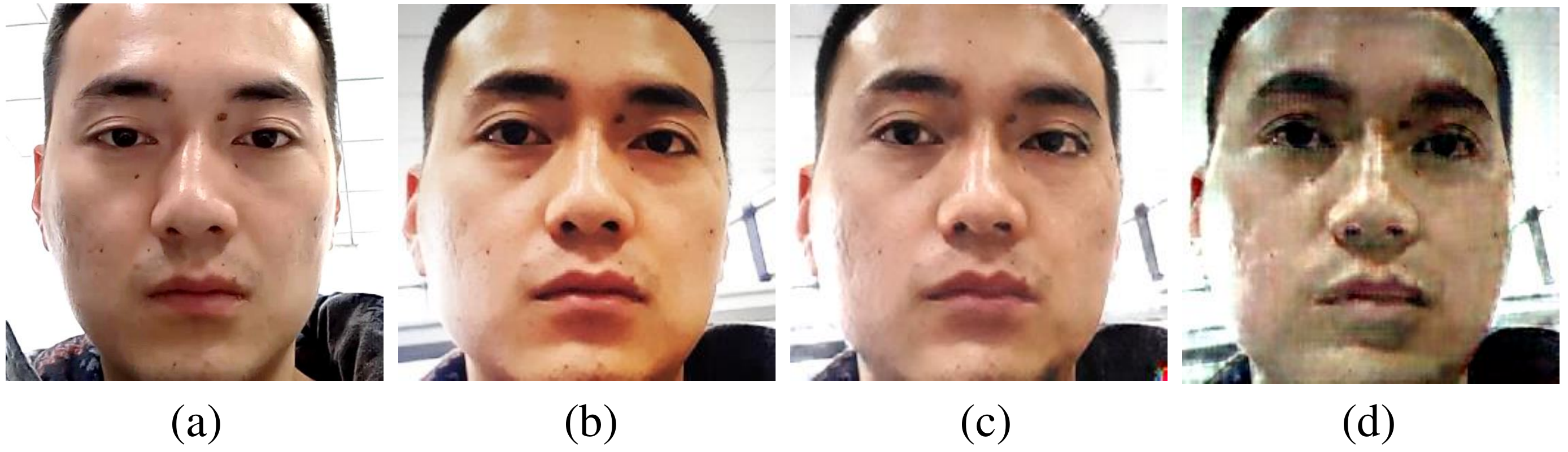}}
}{%
  \caption{\small Live reconstruction comparison: (a) live, (b) spoof, (c) ESR+D-GAN, (d) ESR+GAN.}%
  \label{fig:5}
}
\capbtabbox{%
  \resizebox{0.9\linewidth}{!}{
  \begin{tabular}{llll}
   \toprule
   Method & APCER (\%) & BPCER (\%) & ACER (\%) \\ \midrule
   ESR & $0.8$ & $4.3$ & $2.6$ \\
   ESR+GAN & $1.5$ & $2.7$ & $2.1$ \\
   ESR+D-GAN & $0.8$ & $2.4$ & $1.6$ \\
   ESR+GAN+$L_P$ & $0.8$ & $8.2$ & $4.5$ \\
   ESR+D-GAN+$L_P$ & $\textbf{0.8}$ & $\textbf{1.3}$ & $\textbf{1.1}$ \\ \bottomrule
  \end{tabular}}
}{%
  \caption{\small Quantitative ablation study of components in our approach.}%
  \label{tab:5}
}
\end{floatrow}
\figvspace
\end{figure*}

\SubSection{Spoof Traces Classification}
To quantitatively evaluate the spoof trace disentanglement, we perform a spoof medium classification on the disentangled spoof traces and report the classification accuracy.
The spoof traces should contain spoof medium-specific information, so that they can be used for clustering without seeing the face.
After STDN finishes training with only binary labels, but not the spoof type label, we fix STDN and apply a simple CNN (\textit{i.e.}, AlexNet) on the estimated spoof traces to do a supervised spoof medium classification.
We follow the same testing protocol in~\cite{jourabloo-face-despoofing} in Oulu-NPU Protocol $1$, and the results are shown in Tab.~\ref{tab:cls}. 
Our $3$-class model and $5$-class model can achieve classification accuracy of $92.0\%$ and $83.3\%$ respectively.
Compared with the previous method~\cite{jourabloo-face-despoofing}, we show an improvement of $10\%$ on the $3$-class model and $29\%$ on the $5$-class model.
In addition, we  train the same CNN on the original images instead of the estimated spoof traces for the same spoof medium classification task, and the classification accuracy can only reach $86.3\%$ ($3$-class) and $80.6\%$ ($5$-class).
This further demonstrates that the estimated traces do contain significant information to distinguish different spoof mediums.

\begin{figure*}[t]
\small
\centering
\resizebox{\linewidth}{!}{\includegraphics{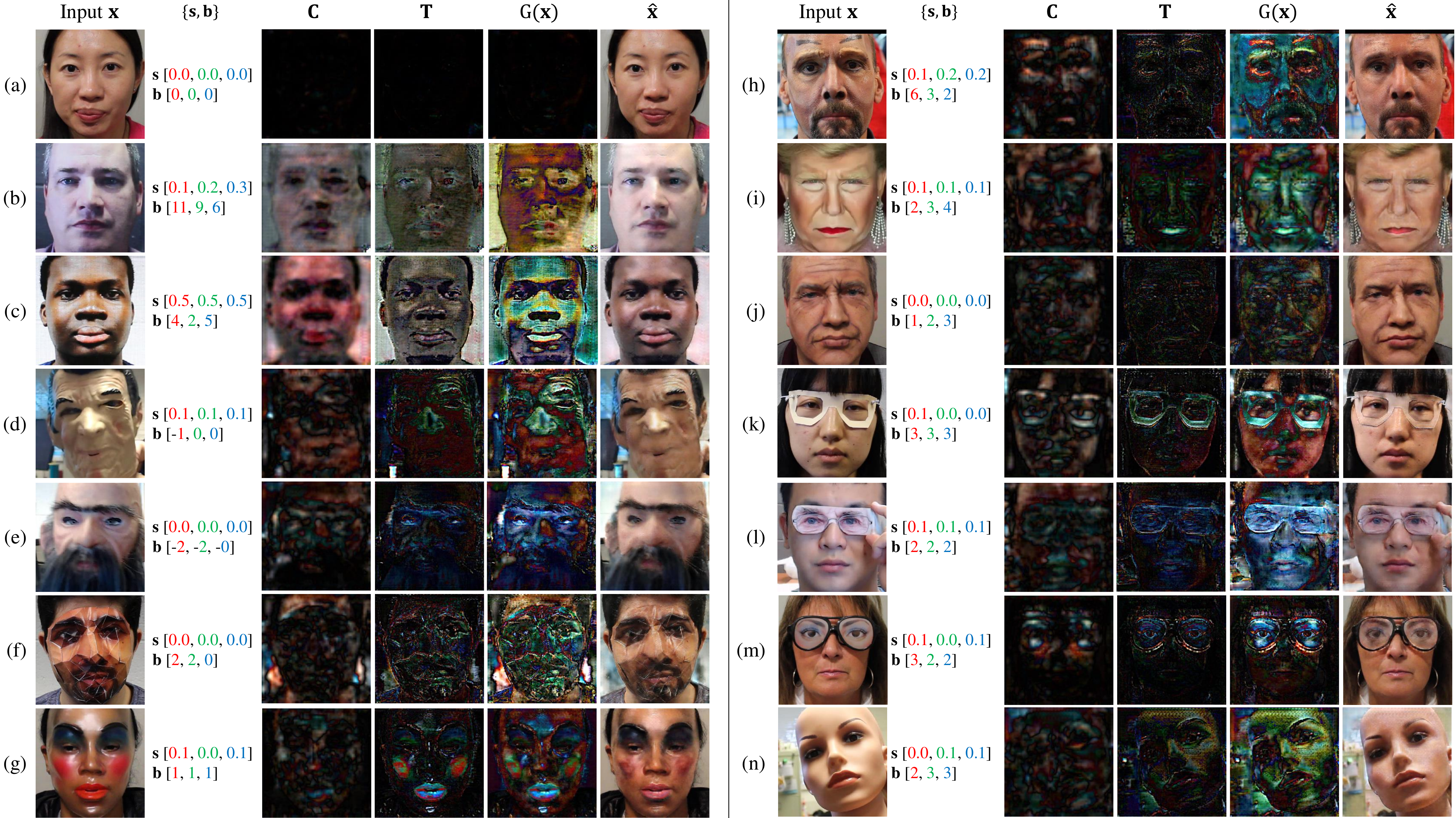}}
\caption{\small Examples of spoof trace disentanglement on SiW-M. The (a)-(n) items are live, print, replay, half mask, silicone mask, paper mask, transparent mask, obfuscation makeup, impersonation makeup, cosmetic makeup, paper glasses, partial paper, funny eye glasses, and mannequin head. The first column is the input face, the $2$nd-$4$th columns are the spoof trace elements $\{\textbf{s},\textbf{b},\textbf{C},\textbf{T}\}$, the $5$th column is the overall spoof traces, and the last column is the reconstructed live.
}
\label{fig:6}
\figvspace
\end{figure*}
\SubSection{Ablation Study}
In this section, we show the importance of each design of our proposed approach on the Oulu-NPU Protocol $1$. 
Our baseline is the encoder with ESR (denoted as ESR), which is a conventional regression model.
To validate the effectiveness of GAN training, we report the results from ESR with GAN. However the generator's output of this model is a single-layer spoof trace with the input size, instead of the proposed four elements.
To demonstrate the effectiveness of disentangled $4$-element spoof trace, we change the single layer to the proposed $\{\textbf{s},\textbf{b},\textbf{C},\textbf{T}\}$, denoted as ESR+D-GAN.
In addition, we evaluate the effect of the training step $3$ via enabling the pixel loss $L_P$ on both ESR+GAN and ESR+D-GAN.
Our final approach is denoted as ESR+D-GAN+$L_P$.

Tab.~\ref{tab:5} shows the results of comparison.
The baseline model can achieve a decent performance of ACER $2.6\%$.
Adding GAN to the baseline can improve the ACER from $2.6\%$ to $2.1\%$, while adding D-GAN can improve to $1.6\%$.
Moreover, ESR+D-GAN can produce spoof traces with much higher quality than ESR+GAN, shown in Fig.~\ref{fig:5}. 
In addition, if the bad-quality spoof samples are used in the training step $3$, it would increase the error rate from $2.1\%$ to $4.5\%$.
On the contrary, when feeding the good-quality synthetic spoof samples to the generator, we can achieve a significant improvement from $1.6\%$ to $1.1\%$, which is the performance of the proposed method.
\SubSection{Visualization}
As shown in Fig.~\ref{fig:6}, we successfully disentangle various spoof traces. 
\textit{E.g.}, strong color distortion shows up in print/replay attacks (Fig.~\ref{fig:6}b-c).
Moiré patterns in the replay attack are well detected (Fig.~\ref{fig:6}c).
For makeup attacks (Fig.~\ref{fig:6}h-j), the fake eyebrows, lipstick, artificial wax, and cheek shade are clearly detected.
The folds and edges in paper-crafted mask (Fig.~\ref{fig:6}f) are well detected.
Although our method cannot provide a convincing estimation for a few spoof types (\textit{e.g.}, funny eye glasses in Fig.~\ref{fig:6}m), the model effectively focuses on the correct region and disentangles parts of the traces. 

Additionally, we show some examples of spoof synthesis using the disentangled spoof traces in Fig.~\ref{fig:7}.
The spoof traces can be precisely transferred to a new face without changing the identity of the target face. 
Thanks to the proposed $3$D warping layer, the geometric discrepancy between the source spoof trace and the target face is corrected during the synthesis.
These two figures demonstrate that our approach disentangles visually convincing spoof traces that help face anti-spoofing.

\Section{Conclusions}
This work proposes a network (STDN) to tackle a challenging problem of disentangling spoof traces from faces.
With the spoof traces, we reconstruct the live faces as well as synthesize new spoofs.
To correct the geometric discrepancy in synthesis, we propose a $3$D warping layer to deform the traces.
The disentanglement not only improves the SOTA of both known and unknown anti-spoofing, but also provides visual evidence to support the model's decision.
\figvspace

\subsubsection{Acknowledgment}
This research is based upon work supported by the Office of the Director of National Intelligence (ODNI), Intelligence Advanced Research Projects Activity (IARPA), via IARPA R\&D Contract No.~$2017$-$17020200004$. The views and conclusions contained herein are those of the authors and should not be interpreted as necessarily representing the official policies or endorsements, either expressed or implied, of the ODNI, IARPA, or the U.S. Government. The U.S. Government is authorized to reproduce and distribute reprints for Governmental purposes notwithstanding any copyright annotation thereon.

\clearpage
\bibliographystyle{splncs04}
\bibliography{bib}
\end{document}